\documentclass[10pt,twocolumn,letterpaper]{article}

\usepackage{cvpr}

\usepackage{graphicx}
\usepackage{booktabs}
\usepackage{xcolor}
\usepackage{graphicx}
\usepackage{subcaption}
\usepackage[dvipsnames]{xcolor}
\usepackage[table]{xcolor}
\usepackage{array}

\usepackage{hyperref}

\hypersetup{
    colorlinks=true,
    urlcolor=magenta, 
    linkcolor=blue,
    citecolor=green
}
\usepackage{colortbl}
\usepackage{booktabs} 
\usepackage{tabularx}
\usepackage{makecell}
\usepackage{booktabs}
\usepackage{pifont}
\usepackage{multirow}
\usepackage{cuted}

\usepackage{capt-of}
\usepackage[accsupp]{axessibility}  

\usepackage{hyperref}

\usepackage{orcidlink}

\usepackage{cuted}
\begin{document}

\title{Benchmarking Vision-Language Models for Microscopic \\ Plant Image Understanding}
\author{
Tianqi Wei$^{1}$,
Xin Yu$^{2}$,
Zhi Chen$^{3}$,
Scott Chapman$^{1}$,
Zi Huang$^{1}$\\
$^{1}$The University of Queensland, 
$^{2}$Adelaide University,
$^{3}$University of Southern Queensland\\
\url{https://github.com/tqwei05/PlantMicro}
}

\maketitle

\begin{strip}
\begin{center}
\vspace{-3.2em}
\includegraphics[width=0.9\textwidth]{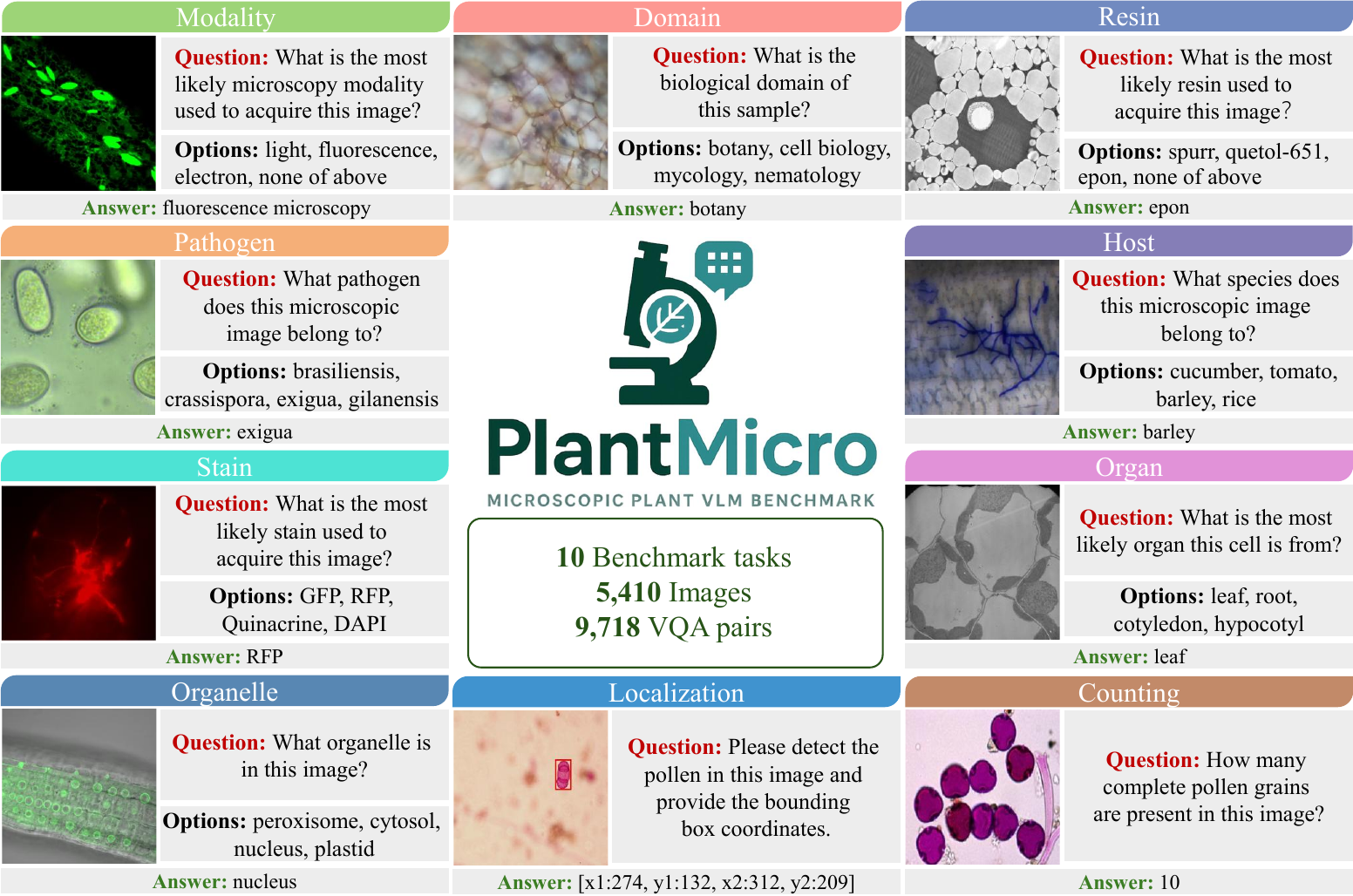}
\captionof{figure}{VQA samples from different tasks in PlantMicro. PlantMicro consists of 5,410 microscopy images and 9,718 VQA pairs designed for microscopic understanding tasks.}
\label{fig:fig1}
\end{center}
\end{strip}

\begin{abstract}
Microscopic imaging provides essential visual evidence for studying plant biology and pathology at the cellular and subcellular levels. However, existing benchmarks for vision-language models primarily focus on macroscopic plant imagery, while the microscopic domain remains underexplored. 
To address this gap, we present PlantMicro, a comprehensive benchmark for evaluating vision–language models (VLMs) in microscopic plant imagery. 
PlantMicro integrates more than 5,000 images collected from diverse hosts, biological domains, and imaging modalities.
Building on this diversity, we design a set of complementary tasks that capture different aspects of microscopic image understanding. To support these tasks, we construct over 9,000 VQA pairs that systematically evaluate the capabilities of VLMs.
Experiments on PlantMicro show that current VLMs struggle with fine-grained recognition and biologically grounded reasoning. For example, GPT-5 achieves 34.93\% accuracy on the pathogen classification task, which is only modestly above a 24.95\% random guessing baseline. The results highlight a significant gap in the ability of current VLMs to comprehend microscopic plant images.
PlantMicro provides a standardized foundation for advancing VLMs toward reliable and comprehensive microscopy-level plant understanding. 
\end{abstract}
\vspace{-1em}

\section{Introduction}
\label{sec:intro}



Microscopy opens a window onto cellular and subcellular structure, revealing tissue organization, cell morphology, and fine biological details that are invisible at the field scale. For plant science, these features underpin explanations of development~\cite{meyerowitz1997plantdev1,ovevcka2018plantdev2,prunet2020plantdev3}, stress adaptation~\cite{lamke2017plantstress1,du2024plantstress2,iqbal2021plantstress3}, and host–pathogen interaction~\cite{rufian2018pathogen1,luck2025barleypowderymidlew}. Yet most evaluations of Vision–Language Models (VLMs) ~\cite{radford2021clip,liu2024llava,team2024gemini,hurst2024gpt,bai2023qwen}  in plants focus on macroscopic scenes \cite{gauba2025agmmu,shinoda2025agrobench,arshad2025ageval,CDDM,wang2025agricm3}. As a result, microscopic visual understanding and its associated domain-specific terminology remain largely unexplored.


Plants are fundamental to life on Earth. They drive primary productivity, release oxygen, and regulate global carbon cycling~\cite{field1998plantintro}. These roles make plant science central to agriculture, ecology, and modern biology, with direct implications for food security and environmental sustainability~\cite{Pawlak2020foodsecurity}. 
With advances in computer vision, plant research has progressed rapidly across tasks such as species classification~\cite{Zhong2019cropcl1,Yordanov2023cropcl2,liu2022plantspeciescls}, disease detection~\cite{Amrani2023aphiddetection,Wei2025augmenttoseg,wei2024plantwild}, and growth-stage recognition~\cite{ni2024stage1,li2025stage2}.


VLMs have emerged as general-purpose learners that link visual content with natural language processing~\cite{radford2021clip,liu2024llava,team2024gemini,hurst2024gpt,bai2023qwen}. Through prompt-based interaction they support open-vocabulary recognition~\cite{radford2021clip}, zero-shot classification~\cite{li2023blip2,zhai2022lit}, and few-shot learning~\cite{alayrac2022flamingo,gao2024clipadapter}. 
Compared with conventional supervised models, VLMs provide an interpretable framework that connects perception with reasoning through language and can perform various tasks without task-specific training.
In plant science, existing VLM benchmarks, \eg, CDDM~\cite{CDDM} and AgroBench~\cite{shinoda2025agrobench}, emphasize field-level imagery and operations such as weed identification, disease recognition, and management assistance, but they largely probe external appearance rather than the internal structures that microscopy exposes.


To bridge this gap, we introduce PlantMicro, a comprehensive benchmark for systematically evaluating VLMs on microscopic plant image understanding. PlantMicro spans multiple biological domains, including mycology, nematology, botany, cell biology, and incorporates various modalities, such as light, fluorescence, and electron microscopy.
We define a diverse set of benchmark tasks spanning attribute prediction, biological entity identification, localization, and counting.
Specifically, image-level attribute tasks include domain, modality, stain, and resin prediction.
Biological entity identification tasks cover host, pathogen, organelle, and organ recognition.
In addition, we introduce localization to assess spatial grounding ability and target counting to evaluate model robustness in microscopic scenarios.
PlantMicro adopts a visual question answering format for evaluation. All questions and annotations are derived from expert-level metadata in the source microscopy datasets and further refined through manual curation to ensure accurate and consistent labeling.
Compared with existing benchmarks that either lack microscopy or do not support VLM benchmarking, PlantMicro allows fine-grained biological interpretation through expert-curated VQA pairs, as summarized in Table~\ref{tab:data_comparison}.
Our main contributions can be summarized as:
\begin{itemize}
    \item We present PlantMicro, a unified benchmark comprising 5,410 plant microscopy images and 9,718 visual question answering (VQA) pairs.
    \item We establish a suite of benchmark tasks derived from the standardized metadata, which provide a systematic evaluation for VLMs across diverse aspects of microscopic plant image understanding.
    \item We benchmark a wide range of VLMs on PlantMicro and provide a comprehensive analysis across diverse microscopic understanding tasks, revealing that VLMs are capable of generic perceptual classification tasks but struggle with fine-grained biological recognition, counting and localization tasks.
\end{itemize}

\begin{table*}[t]
\centering
\caption{Comparison of image benchmarks for plant science.}
\label{tab:data_comparison}
\begin{tabular}{l c c c c c c c c}
\toprule
\multirow{2}{*}{\textbf{Benchmarks}}
& \multirow{2}{*}{\textbf{\#Host}}
& \multirow{2}{*}{\textbf{\#Images}}
& \multirow{2}{*}{\textbf{\#VQA Pairs}}
& \multirow{2}{*}{\textbf{Expert}}
& \multirow{2}{*}{\textbf{Micro.}}
& \multicolumn{3}{c}{\textbf{Imaging Modality}} \\
\cmidrule(lr){7-9}
& & & & & & \textbf{Light} & \textbf{Fluor.} & \textbf{Electron} \\
\midrule
Tomato Spore~\cite{javidan2024tomatospore} & 1 & 400 & -- & \ding{51} & \ding{51} & \ding{51} & -- & -- \\
PODB~\cite{mano2007PODB} & 13 & 459 & -- & \ding{51} & \ding{51} & -- & \ding{51} & -- \\
Hydrous Plants~\cite{takehara2020nanosuit} & 13 & 218 & -- & \ding{51} & \ding{51} & -- & -- & \ding{51} \\
CDDM~\cite{CDDM} & 15 & 137k & 1M & -- & -- & -- & -- & -- \\
AgroBench~\cite{shinoda2025agrobench} & 311 & 3,745 & 4,342 & \ding{51} & -- & -- & -- & -- \\
Agri-CM$^3$~\cite{wang2025agricm3} & 11 & 3,939 & 15,901 & \ding{51} & -- & -- & -- & -- \\
\midrule
\textbf{PlantMicro (ours)} & 25 & 5,410 & 9,718 & \ding{51} & \ding{51} & \ding{51} & \ding{51} & \ding{51} \\
\bottomrule
\end{tabular}
\end{table*}

\section{Related Work}
\label{sec:related_work}
\noindent \textbf{Computer Vision in Plant Science.}
Computer vision has been widely applied in plant science for diverse visual perception and analysis tasks.
Early studies focused on image classification, recognizing plant species or diseases~\cite{liu2022plantspeciesidentification,wang2021tcnn,wei2024snap}. Representative classification datasets like PlantVillage~\cite{mohanty2016plantvillage} provide clean, laboratory-collected images for disease identification, while PlantDoc~\cite{singh2020plantdoc} and PlantWild~\cite{wei2024plantwild} offer in-the-wild samples with complex backgrounds and varying lighting conditions.
Subsequent works extended to object detection, where deep models such as Faster R-CNN~\cite{ren2015fasterrcnn} and YOLO~\cite{redmon2016yolo} enable localization and quantification of objects such as fruits, weeds, and insects in complex field environments~\cite{lim2024trackpepper,lu2023odlfruitlocalization,wang2019review_weed_detection,chakrabarty2024insectdetection}.
Segmentation methods further advance fine-grained understanding by predicting pixel-level regions for detailed analysis. These methods have been applied to diverse tasks, including leaf and canopy analysis, spike segmentation in cereals, and disease lesion delineation~\cite{gong2024segmentationmaize,wang2025gwfss,wei2024plantseg,zhang2022wheatnet}.
Although traditional methods have advanced research in plant imagery, they still rely on dedicated datasets, annotations, and model architectures, which limit scalability and generalization under different tasks.

\noindent \textbf{Field-Level Plant Benchmarks for VLMs.}
Conventional vision models~\cite{Zhong2019cropcl1,liu2022plantspeciescls,Amrani2023aphiddetection,Wei2025augmenttoseg} are task-specific and depend heavily on dedicated annotations, which limit their scalability and generalization. Vision–language models (VLMs)~\cite{radford2021clip,bai2023qwen,comanici2025gemini2.5,gpt4omini} are trained on massive-scale multimodal data to align visual and textual information, enabling open-vocabulary recognition and reasoning.
Inspired by their success in general domains, recent benchmarks have begun extending VLM evaluation to field-level plant imagery.
To diagnose plant diseases, CDDM~\cite{CDDM} constructed a large-scale multimodal benchmark with over 130k plant images and one million question–answer pairs.
AgEval~\cite{arshad2025ageval} covers twelve plant stress tasks under zero and few-shot settings, indicating that model performance depends strongly on how prompts are designed and whether the provided examples match the target domain.
AgroBench~\cite{shinoda2025agrobench} offers a large-scale evaluation across seven agricultural themes and finds that current VLMs still struggle with fine-grained recognition tasks such as weed and disease identification.
Agri-CM$^3$~\cite{wang2025agricm3} proposes a hierarchical evaluation framework that assesses plant understanding across progressively complex levels of perception, reasoning, and knowledge application.
MIRAGE~\cite{dongre2025mirage} focuses on expert-guided multimodal conversations, where models need to interpret ambiguous inputs and produce practical responses.
However, existing VLM benchmarks focus mostly on macroscopic plant imagery, leaving microscopic plant understanding largely underexplored.

\noindent \textbf{Microscopic Benchmarks for VLMs.}
Recent work has begun extending VLMs into microscopic and biomedical domains.
BiomedCLIP~\cite{zhang2023biomedclip}, pretrained on 15 million biomedical image–text pairs, provides a strong foundation for fine-grained tasks such as pneumonia detection.
PMC-CLIP~\cite{lin2023pmcclip} leverages large-scale figure–caption data from biomedical literature and achieves strong performance in medical visual question answering.
PLIP~\cite{huang2023plip}, trained on pathology images and text from Medical Twitter, learns diverse pathological patterns and diagnostic expressions from real-world data.
Benchmarks such as MicroBench~\cite{lozanomicrobench} have also been introduced to evaluate VLMs across microscopy modalities and biomedical tasks, covering light, fluorescence, and electron imaging for classification, captioning, and question answering.
Despite these advances, most existing efforts primarily focus on human or animal biomedical imagery. In contrast, plant microscopy, with its distinct cellular organization and biological processes, remains an underexplored area. To fill this gap, we develop a unified framework for evaluating VLMs in microscopic plant understanding, bridging visual perception with plant biological reasoning.

\section{The PlantMicro Benchmark}
\label{sec:Methodology}
\begin{figure*}[t]
    \centering
    \includegraphics[width=\textwidth,height=6cm,keepaspectratio]{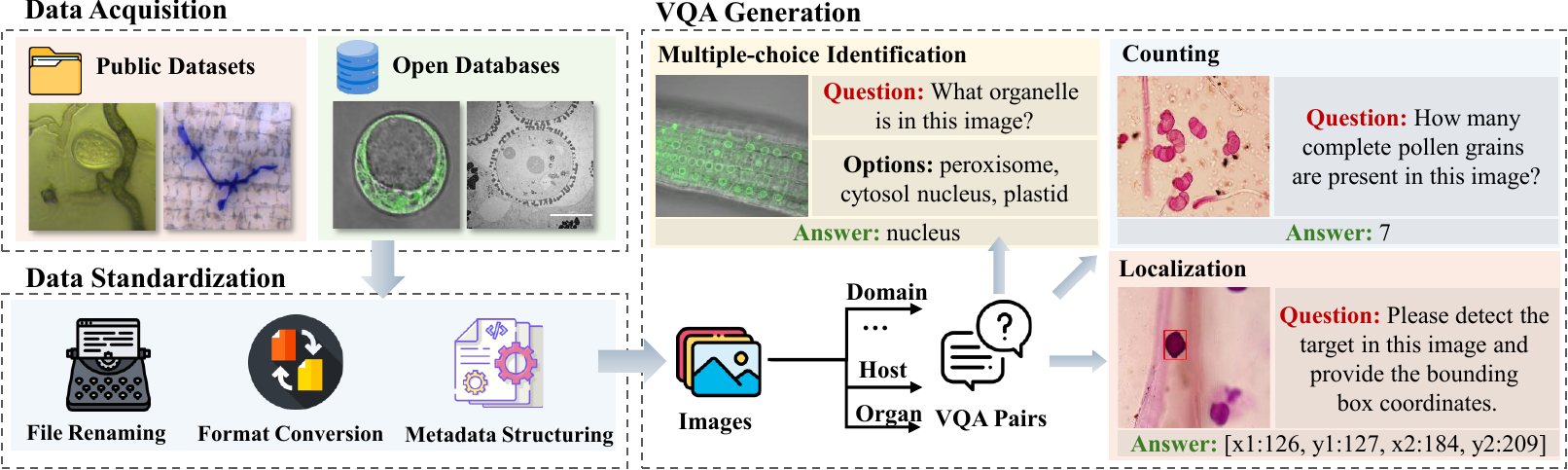}
    \vspace{-1.5em}
    \caption{Construction pipeline of the PlantMicro benchmark. The process includes data acquisition from public data sources, followed by standardized curation and metadata verification. Based on the validated attributes, we generate structured VQA pairs for multiple-choice identification, object localization, and target counting tasks.}
    \label{fig:curation}
    \vspace{-0em}
\end{figure*}

\begin{figure*}[t]
    \centering
    \begin{subfigure}[t]{0.27\linewidth}
        \centering
        \includegraphics[width=\linewidth]{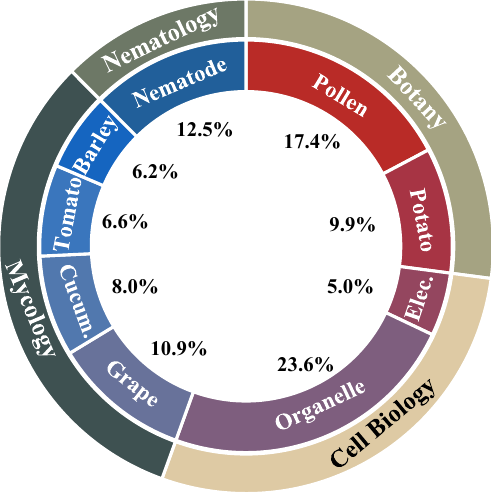}
        \caption{Domain distribution}
        \label{fig:crop_domain}
    \end{subfigure}
    \begin{subfigure}[t]{0.27\linewidth}
        \centering
        \includegraphics[width=\linewidth]{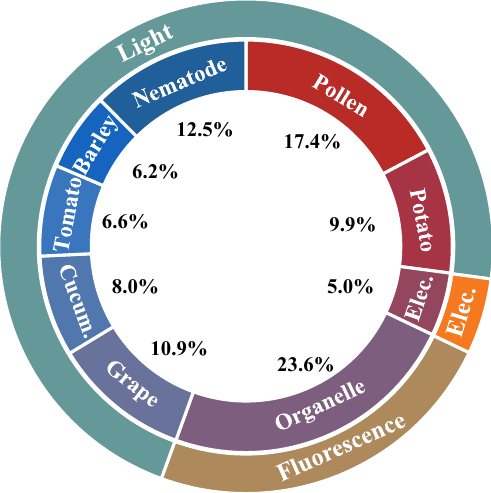}
        \caption{Modality distribution}
        \label{fig:crop_modality}
    \end{subfigure}
    \begin{subfigure}[t]{0.35\linewidth}
        \centering
        \includegraphics[width=\linewidth]{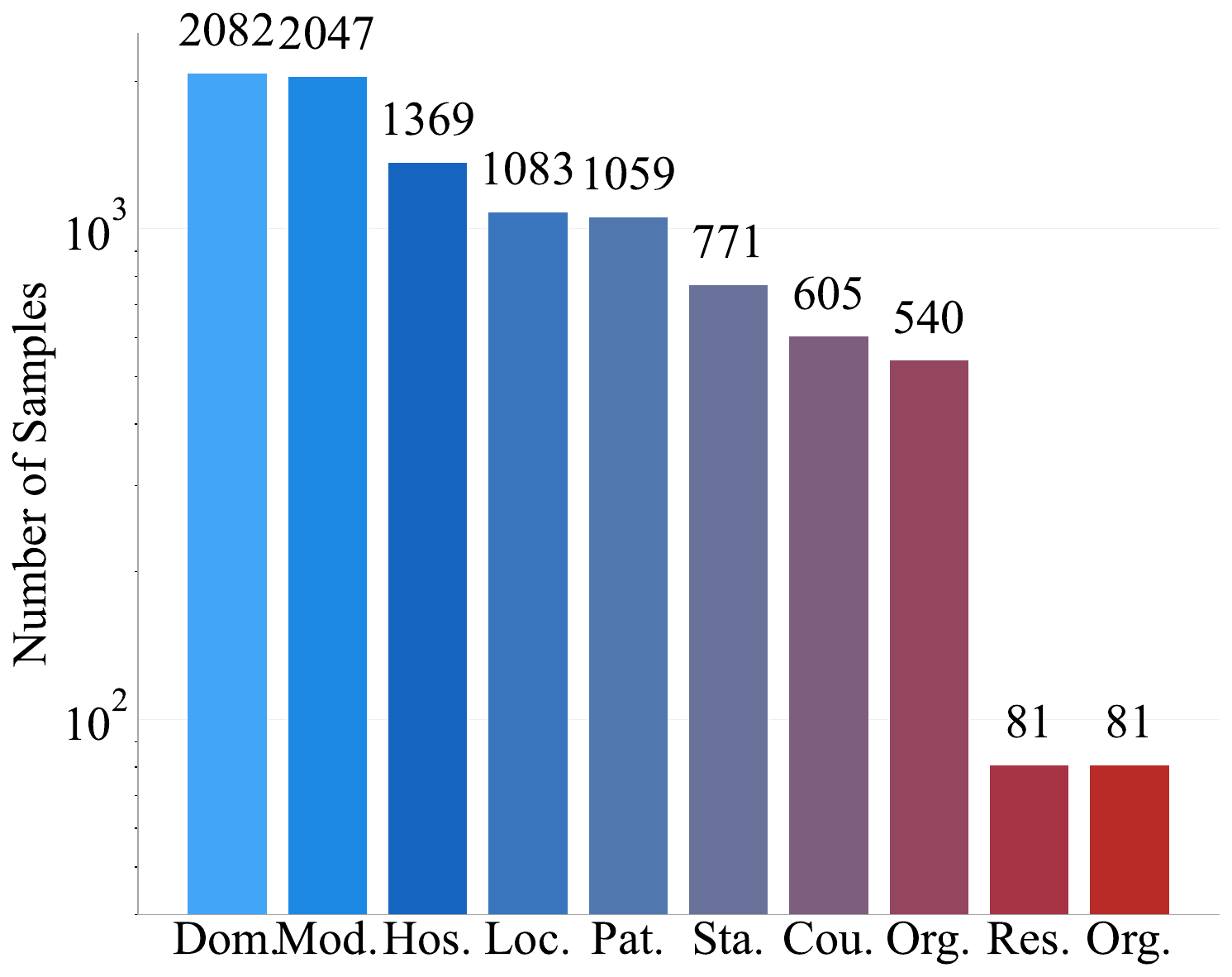}
        \caption{VQA task statistics}
        \label{fig:VQA_tasks}
    \end{subfigure}
    \vspace{-0.5em}
    \caption{Detailed statistics of \textbf{PlantMicro}. (a) Domain distribution. (b) Modality distribution. (c) The number of QA pairs of each benchmark task.} 
    \vspace{-0.5em}
    \label{fig:dataset_stats}
\end{figure*}

PlantMicro is a comprehensive benchmark for evaluating vision–language models (VLMs) on microscopic plant image understanding. 
We illustrate the construction process of our PlantMicro in Figure~\ref{fig:curation}.
It aggregates diverse microscopy images collected from public datasets in peer-reviewed publications and institutional repositories. The scope includes plant pathology, botany, cell biology, and nematology. In addition, it covers a broad range of biological contexts and imaging modalities.
We carefully curated, standardized, and annotated all images with unified metadata. Thus, the benchmark allows consistent evaluation across a wide range of tasks. 
In total, PlantMicro contains 9,718 VQA pairs derived from 5,410 microscopy images. Figure~\ref{fig:crop_domain} and \ref{fig:crop_modality} illustrate the distribution of PlantMicro across biological domains and microscopy modalities, respectively. Figure~\ref{fig:VQA_tasks} reports the number of VQA pairs for each benchmark task.

\subsection{Data Review and Acquisition}
We conduct a systematic search on peer-reviewed publications~\cite{zhu2024cucumberspore,javidan2024tomatospore,luck2025barleypowderymidlew,crespo2023grapespore,indarti2025nematode,biswas2020potatotuber,zu2024pollen}, and open data repositories~\cite{mano2007PODB,mano2014electronPODB} to collect microscopy datasets of plant tissues and diseases. 
To ensure data diversity, we include multiple biological domains, such as mycology, botany, and nematology, as well as imaging modalities, including light, fluorescence, and electron microscopy.
We prioritize datasets that are openly available for research use. For those not directly accessible, we contact the corresponding authors to obtain permission for research use.

\subsection{Data Standardization} 

The selected datasets differ in file organization, annotation formats, and metadata completeness. To ensure structural consistency, we standardize all data under a unified curation protocol. Original images stored in heterogeneous formats such as TIFF, JPEG, and PNG, with inconsistent naming conventions, are systematically reorganized. To prevent filename conflicts across sources, each image is assigned a unique identifier while preserving its source information. All images are converted to JPEG format at their original resolution. For each image, we construct a structured JSON record to store curated metadata.
Key attributes such as biological domain, imaging modality, staining method, and host species are verified through original publications, communication with dataset providers, or consultation with domain experts. To further ensure reliability, all attributes are independently reviewed by two biologists, and samples with inconsistent judgments are removed.


\subsection{Benchmark Tasks} 
Building on the established metadata, we introduce a suite of benchmark tasks to evaluate VLMs across different aspects of microscopic plant image understanding.
The tasks are introduced as follows.

\noindent \textbf{1) Modality Identification.} 
This task focuses on identifying the microscopy modality used to acquire each image. PlantMicro includes three primary microscopy modalities: light microscopy, fluorescence microscopy, and electron microscopy. This task evaluates VLMs’ ability to recognize modality-specific visual cues, such as bright-field illumination patterns in light microscopy, fluorescent signals in fluorescence microscopy, and fine surface textures in electron microscopy.

\noindent \textbf{2) Domain Classification.} 
This task aims to identify the biological domain of each image based on its visual characteristics. The images in PlantMicro are grouped into four major domains: Mycology, Nematology, Botany, and Cell Biology. The four domains correspond to fungal, nematode, plant tissue, and cellular imaging studies. This task evaluates a VLM’s ability to understand different biological contexts in order to recognize domain-specific visual patterns such as spore morphology, nematode anatomy, or cellular structures.

\noindent \textbf{3) Resin Identification.} 
This task aims to identify the embedding resin used in electron microscopy sample preparation. 
Resin is used to embed and support biological samples so that they can be imaged and preserved without distortion, especially for electron microscopy and high-resolution light microscopy. Different resins provide varying levels of mechanical support and structural stability, which may influence image contrast, sharpness, and background consistency. PlantMicro includes samples embedded with commonly used resins such as Epon, Spurr, and Quetol-651. These materials lead to differences in imaging appearance, including contrast clarity, and and background uniformity. This task evaluates whether VLMs can capture these visual cues associated with different resins.

\noindent \textbf{4) Stain/Dye Recognition.} 
Stains and dyes are used to add contrast and specificity so microscopic structures become distinguishable.
This task aims to identify the staining or fluorescent dye applied in the microscopic imaging process. PlantMicro contains samples labeled with various dyes such as DAPI, Green Fluorescent Protein (GFP), and Red Fluorescent Protein (RFP), which include both chemical and fluorescence staining. Different dyes result in distinct colors and signal distributions. This task evaluates the VLMs' ability to understand staining techniques and biological semantics purely from visual appearance.

\noindent \textbf{5) Host Identification.} 
In this task, VLMs are required to recognize the host organism shown in each microscopic image. The host refers to the plant species from which the sample is taken. Unlike macroscopic photos, microscopic images only show local cell or tissue structures without clear morphological contexts, making recognition difficult. Models need to use fine visual cues such as cell wall patterns, organelle arrangement, and color variations to infer which plant species the sample belongs to.

\noindent \textbf{6) Pathogen Classification.} 
This task involves fine-grained classification within each biological domain. In the fungal datasets, the goal is to distinguish different pathogenic species or infection types, while in the nematode datasets, it focuses on recognizing individual nematode species. These categories often exhibit high intra-class similarity and subtle inter-class differences in texture, shape, and structural patterns. The task evaluates whether models can capture fine-grained visual features to achieve accurate recognition within each domain.


\noindent \textbf{7) Organelle Detection.} 
The task assesses whether VLMs can robustly detect subcellular structures under heterogeneous visual conditions. Common targets include nuclei, mitochondria, plastids, and vacuoles. These structures differ in shape, texture, and contrast, but their appearances can vary across cell types, stains, and imaging modalities. The evaluation measures VLMs’ capability to recognize organelles in microscopic images.

\noindent \textbf{8) Organ Prediction.} 
This task predicts which plant organ a microscopic image originates from, for example, leaf, root, cotyledon, or seed. It examines whether VLMs can relate cellular organization and tissue morphology to organ-level identity, capturing the structural hierarchy from microscopic features to macroscopic anatomy.

\noindent \textbf{9) Target Counting.} 
This task evaluates whether VLMs can accurately enumerate microscopic entities, such as spores or pollen grains, in densely populated visual scenes. Counting requires distinguishing true targets from visually similar debris and background artifacts, as well as separating tightly clustered instances, testing VLMs' ability for fine-grained visual parsing and quantitative reasoning. The average of objects per image is 8.2, and the median is 6.

\noindent \textbf{10) Object Localization.} 
This task evaluates the spatial grounding ability of VLMs by localizing a specified target and the prediction of its bounding box coordinates (\ie, absolute pixel coordinates $(x_1, y_1, x_2, y_2)$).

Tasks 1-8 are evaluated under a multiple-choice setting, where models select the correct answer from four candidate options. For Tasks 9 and 10, we additionally evaluate models under a direct prediction setting, where models directly generate the target count or bounding box coordinates in a predefined format. Predictions that violate the required format are treated as failures.



\begin{table*}[t]
\centering
\caption{Benchmark performance of vision–language models on PlantMicro. Results are provided for closed-source and open-source VLMs evaluated on identification tasks.}
\begin{tabular}{l|cccc|cccc|cc}
\toprule
\multirow{2}{*}{\textbf{Model}} &
\multicolumn{4}{c|}{\textbf{Image-Level}} &
\multicolumn{4}{c|}{\textbf{Biological-Level}} &
\multicolumn{2}{c}{\textbf{Overall}} \\
\cmidrule(lr){2-5}
\cmidrule(lr){6-9}
\cmidrule(lr){10-11}
& \textbf{Mod.} & \textbf{Dom.} & \textbf{Res.} & \textbf{Sta.}
& \textbf{Host} & \textbf{Path.} & \textbf{OrgL.} & \textbf{Org.}
& \textbf{Macro} & \textbf{Micro} \\
\midrule
Random Choice 
& 24.97 & 25.02 & 25.21 & 25.52 
& 24.95 & 24.89 & 24.99 & 23.90 
& 24.93 & 25.23 \\
\midrule
\multicolumn{11}{c}{\textbf{Closed-Source Vision Language Models(VLMs)}} \\
\midrule
GPT-4o mini~\cite{gpt4omini} 
& 73.23 & 62.97 & 51.85 & 50.03 
& 30.78 & 29.00 & 49.26 & 62.96 
& 51.26 & 53.34 \\

GPT-5 mini~\cite{gpt5mini} 
& \textbf{99.73} & 57.54 & 50.62 & 52.51 
& 31.40 & \textbf{43.91} & 50.74 & 71.60 
& 57.26 & 61.17 \\

GPT-5~\cite{OpenAI2025GPT5} 
& 97.72 & 55.46 & 51.85 & 58.46 
& 34.93 & 42.71 & 49.07 & 64.19 
& 56.79 & 60.56 \\

Gemini 2.5-Flash-lite~\cite{comanici2025gemini2.5} 
& 97.51 & 57.44 & 60.49 & 56.80 
& 33.60 & 33.05 & \textbf{56.48} & 65.43 
& 57.60 & 60.36 \\

Gemini 2.5-Flash~\cite{comanici2025gemini2.5} 
& 97.86 & 62.54 & \textbf{85.19} & \textbf{61.45} 
& 45.37 & 32.67 & 51.67 & \textbf{67.90} 
& \textbf{63.08} & 64.12 \\

Gemini 2.5-Pro~\cite{comanici2025gemini2.5} 
& 99.56 & \textbf{63.86} & 80.25 & 59.02 
& \textbf{46.82} & 33.19 & 53.89 & 66.91 
& 62.93 & \textbf{65.07} \\

\rowcolor{gray!20}\textbf{Avg (per task)} 
& 94.27 & 59.97 & 63.38 & 56.38
& 37.15 & 35.76 & 51.85 & 66.50
& 58.15 & 60.77 \\

\midrule

\multicolumn{11}{c}{\textbf{Open-Source Vision Language Models (VLMs)}} \\
\midrule
CLIP-ViT/32~\cite{radford2021clip} 
& 67.37 & 33.05 & 9.64 & 48.51 
& 31.70 & \textbf{38.24} & 26.85 & 16.05 
& 33.93 & 42.90 \\

CLIP-ViT/16~\cite{radford2021clip} 
& 83.10 & 29.68 & 13.58 & 42.54 
& 30.75 & 26.82 & 29.67 & 19.75 
& 34.49 & 44.07 \\

LLaVA-7B~\cite{liu2024llava} 
& 30.39 & 32.47 & 30.86 & 38.91 
& 23.81 & 25.12 & 35.37 & 20.99 
& 29.74 & 30.18 \\

LLaVA-13B~\cite{liu2024llava} 
& 55.50 & 44.68 & 35.80 & 39.30 
& 32.43 & 28.32 & 37.22 & 24.69 
& 37.24 & 41.88 \\

LLaVA-Next-7B~\cite{li2024llavanext} 
& 29.11 & 34.49 & 32.10 & 50.45 
& 23.08 & 19.55 & 30.19 & 49.38 
& 33.54 & 30.57 \\

LLaVA-Next-13B~\cite{li2024llavanext} 
& 49.15 & 32.56 & 49.38 & \textbf{49.97} 
& \textbf{40.10} & {29.56} & 32.04 & 46.91 
& 41.20 & 39.63 \\

QwenVLM-7B~\cite{bai2023qwen} 
& \textbf{68.05} & \textbf{52.11} & \textbf{82.72} & 49.68 
& 30.75 & 25.21 & \textbf{43.15} & \textbf{64.20} 
& \textbf{51.98} & \textbf{48.58} \\

\rowcolor{gray!20}\textbf{Avg (per task)} 
& 54.67 & 37.01 & 36.30 & 45.62
& 30.37 & 27.55 & 33.50 & 34.57
& 37.45 & 39.69 \\




\bottomrule
\end{tabular}

\label{tab:main_results}
\end{table*}

\subsection{VQA Generation} 
Based on the defined benchmark tasks for PlantMicro, we construct a large collection of closed visual question answering (VQA) pairs to evaluate VLM performance across multiple aspects of microscopic image understanding. 
Each VQA pair follows a multiple-choice format, where models are required to select the correct answer from a set of candidate options.
The questions and the ground truth answers are derived directly from the metadata associated with the microscopy images. For recognition tasks, distractor options are sampled from other valid values within the same metadata field to ensure semantic consistency and biological plausibility while avoiding visually implausible choices. For the counting task, distractor options are generated by perturbing the ground-truth count with nearby integer values to produce numerically plausible alternatives. For the localization task in the multiple-choice setting, candidate bounding boxes include the ground-truth box and several spatially perturbed alternatives with similar sizes but shifted positions. The candidate boxes are designed to be non-overlapping to ensure clear spatial separation among options.
Since source microscopy datasets provide heterogeneous metadata fields (\eg, some include ``organelle'' annotations while others do not, and some images contain multiple objects whereas others contain only one), the available question types differ across sources. Accordingly, the VQA pairs are generated by aligning each dataset with its valid metadata fields, ensuring that every task is derived only from images containing the required information. 
All generated VQA pairs are independently reviewed by two domain experts with backgrounds in plant pathology and microscopy to ensure reliability. The experts examine the correctness of the attribute–question alignment, the consistency between visual evidence and the ground-truth answer, and the biological plausibility of all candidate options. Samples with inconsistent judgments are excluded.




\section{Experiments}
\subsection{Experiment Settings}


\noindent \textbf{Benchmarking VLMs.} We evaluate a diverse set of VLMs that differ in architecture and size on the PlantMicro benchmark. 
These VLMs fall into two categories: closed-source models, including the GPT series~\cite{gpt5mini}, Gemini series~\cite{comanici2025gemini2.5}, and open-source models, including CLIP~\cite{radford2021clip}, LLaVA~\cite{liu2024llava}, LLaVA-Next~\cite{li2024llavanext}, and QwenVLM~\cite{bai2023qwen}.
To ensure a fair comparison, all models are assessed under a unified evaluation pipeline that standardizes input processing and prompt design. We conduct experiments on a workstation with two NVIDIA A6000 GPUs.

\noindent \textbf{Evaluation Metrics.}
We evaluate VLMs across three task categories with task-specific metrics.
For multiple-choice identification tasks, we report both macro and micro accuracies. 
The macro and micro accuracy are defined as 
$\text{Micro}=\sum_{i=1}^{C}TP_i/\sum_{i=1}^{C}N_i$, 
and
$\text{Macro}=\frac{1}{C}\sum_{i=1}^{C}(TP_i/N_i)$,
where $C$ is the number of classes, $TP_i$ is the number of correctly predicted samples for class $i$, and $N_i$ is the total number of samples for class $i$.
For counting tasks, models predict the number of target instances per image, and we report Mean Absolute Error (MAE) and Root Mean Square Error (RMSE) to quantify numerical deviation from ground truth.
For localization tasks, models output bounding box coordinates, and performance is evaluated using mean Average Precision at an IoU (Intersection over Union) threshold of 0.5 (mAP@0.5) together with average IoU.
To ensure consistent evaluation, all models are prompted with task-specific formats, including multiple-choice selection for identification, integer-only outputs for counting, and fixed-format bounding box predictions for localization. The predictions are regarded as incorrect if they violate the required format or differ from the ground-truth annotations.

\begin{table*}[t]
\centering
\caption{Performance comparison on counting and localization tasks. 
Evaluation is conducted under two settings: multiple-choice selection and direct output prediction.}
\label{tab:regression_results}

\begin{tabular}{l|c|cc|c|cc}
\toprule
\multirow{2}{*}{\textbf{Model}} 
& \multicolumn{3}{c|}{\textbf{Counting}}
& \multicolumn{3}{c}{\textbf{Localization}} \\

\cmidrule(lr){2-5} \cmidrule(lr){5-7}
& \textbf{MC Acc}$\uparrow$
& \textbf{MAE}$\downarrow$ 
& \textbf{RMSE}$\downarrow$
& \textbf{MC Acc}$\uparrow$
& \textbf{mAP@0.5}$\uparrow$
& \textbf{IoU}$\uparrow$ \\

\midrule

GPT-5-mini~\cite{gpt5mini} & 66.29 & 2.84 & 3.70 & 79.33 & 23.64 & 30.91 \\
GPT-5~\cite{OpenAI2025GPT5} & \textbf{71.34} & \textbf{2.07} & \textbf{3.02} & 83.21 & 56.66 & 50.53 \\
Gemini 2.5-Flash~\cite{comanici2025gemini2.5} & 40.62 & 3.39 & 7.98 & 62.77 & 27.01 & 34.16 \\
Gemini 2.5-Pro~\cite{comanici2025gemini2.5} & 41.37 & 3.30 & 6.32 & 65.91 & 27.60 & 37.75 \\

\midrule

LLaVA-7B~\cite{liu2024llava} & 27.94& 11.95 & 27.81 & 28.44 & 23.54 & 31.70 \\
LLaVA-13B~\cite{liu2024llava} & 28.10& 8.71 & 22.18 & 32.41 & 30.78 & 40.33 \\
LLaVA-Next-7B~\cite{li2024llavanext} & 28.76& 8.67 & 22.75 & 31.98 & 28.10 & 34.48 \\
LLaVA-Next-13B~\cite{li2024llavanext} & 35.37& 5.85 & 13.76 & 35.02 & 35.50 & 42.17 \\
QwenVLM-7B~\cite{bai2023qwen} & 46.45 & 3.38 & 9.24 & \textbf{89.44} & \textbf{65.05} & \textbf{68.64} \\

\bottomrule
\end{tabular}
\end{table*}

\subsection{Main Results}
\noindent \textbf{Multiple-choice Recognition Performance Analysis.}
Table~\ref{tab:main_results} summarizes the main results, with the last row reporting the overall performance of all VLMs across recognition tasks. 
Performance variation across tasks reflects inherent differences in task difficulty. While modality, domain, resin, and stain identification largely rely on global visual cues, the remaining tasks demand fine-grained recognition of structurally diverse microscopic entities.
VLMs achieve their strongest performance on the modality identification task. State-of-the-art closed-source models, such as GPT-5 and Gemini 2.5, surpass 95\% accuracy, whereas open-source counterparts trail by a substantial margin, highlighting a pronounced capability gap in capturing global visual patterns.
Both closed and open-source VLMs perform poorly on host and pathogen identification, with average accuracies around 30\%. These results are marginally above random guessing, which suggests that VLMs still face considerable challenges in identifying host or pathogen types from microscopic evidence. 
For the remaining tasks, the average accuracies generally fall within the 40–50\% range, among which stain recognition performs best. This may be attributed to the ability of VLMs to associate distinct color cues with specific stains, such as green fluorescence from GFP (green fluorescent protein).
Overall, the results across tasks reveal that while VLMs are capable of coarse perceptual discrimination, they still fall short in fine-grained and biologically grounded understanding.

\begin{figure}[t]
    \centering
    \includegraphics[width=0.95\linewidth]{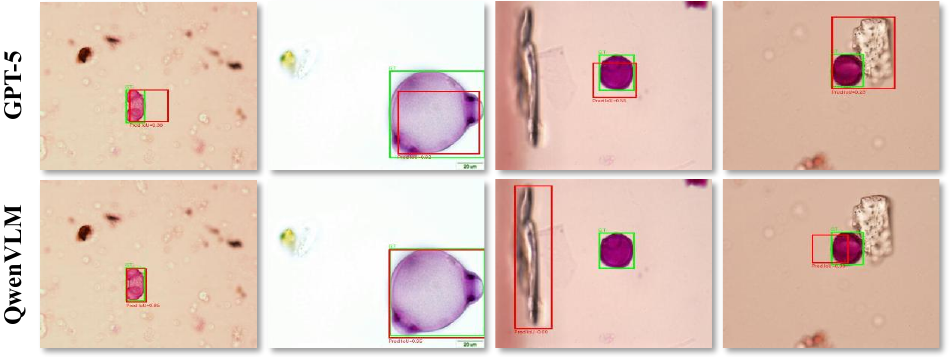}
    \caption{Localization samples of GPT-5 and QwenVLM-7B. Red boxes denote model predictions, while green boxes indicate ground-truth bounding boxes.}
    \label{fig:loc_vis}
\end{figure}

\noindent \textbf{VLM Performance Comparison.}
Across most benchmark tasks, closed-source models consistently outperform open-source counterparts. Among them, Gemini 2.5-Flash achieves the highest overall performance, reaching a macro accuracy of 63.08\%, highlighting the advantage of large proprietary models in visual–textual alignment under microscopic settings. Interestingly, GPT-5-mini slightly surpasses GPT-5 on several tasks, suggesting that smaller models may exhibit stronger robustness in certain fine-grained microscopic contexts.
Open-source VLMs show greater performance variability. QwenVLM-7B attains the highest macro accuracy within this group (51.98\%), approaching the lower range of closed-source performance. Models based on the LLaVA architecture show moderate performance, struggling with pathogen and host recognition. CLIP models perform substantially worse, particularly on resin and organ prediction tasks, where accuracies fall below the random baseline. This performance gap may be related to their contrastive pretraining on natural images, which limits adaptation to the visual patterns of microscopic domains.
Overall, a consistent gap exceeding 10 percentage points remains between closed- and open-source models in both macro and micro accuracy, indicating that current open-source VLMs still face challenges when generalizing to specialized microscopic plant imagery without domain-specific adaptation.

\begin{figure}[t]
    \centering
    \begin{subfigure}[t]{0.49\linewidth}
        \centering
        \includegraphics[height=3.5cm,keepaspectratio]{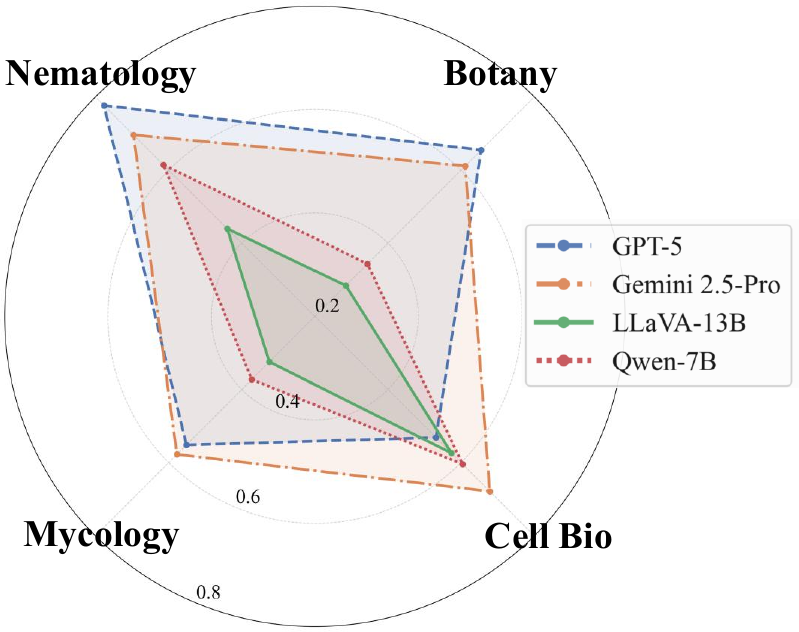}
        \caption{Domain-wise comparison}
        \label{fig:domain}
    \end{subfigure}
    \begin{subfigure}[t]{0.49\linewidth}
        \centering
        \includegraphics[height=3.5cm,keepaspectratio]{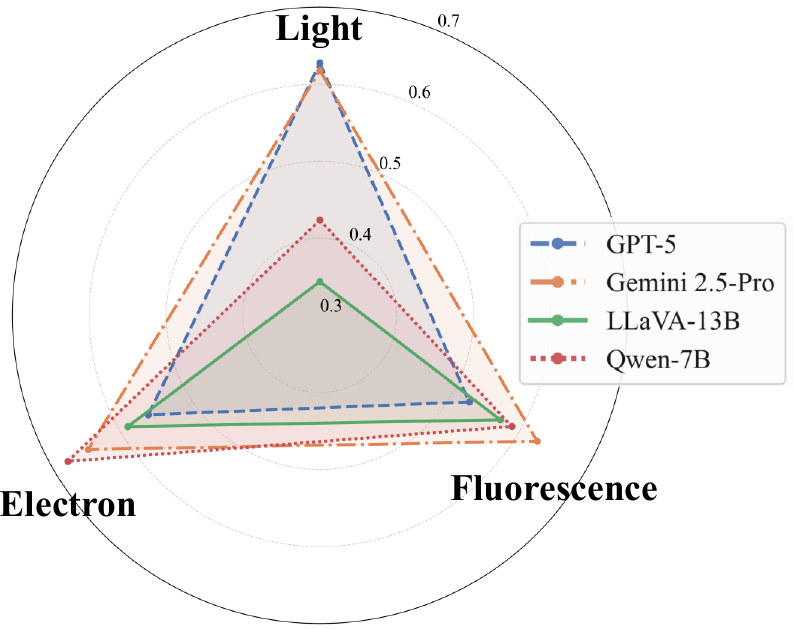}
        \caption{Modality-wise comparison}
        \label{fig:modality}
    \end{subfigure}
    \caption{Accuracy comparison across biological domains and imaging modalities.}
    \label{fig:modality_comparison}
\end{figure}

\begin{table*}[t]
\centering
\caption{Effect of CoT prompting on Gemini 2.5-Pro and QwenVLM-7B.}
\begin{tabular}{l| c| cccc| cccc| cc}
\toprule
\multirow{2}{*}{\textbf{Model}} & 
\multirow{2}{*}{\textbf{CoT}} &
\multicolumn{4}{c|}{\textbf{Image-Level}} &
\multicolumn{4}{c|}{\textbf{Biological-Level}} &
\multicolumn{2}{c}{\textbf{Overall}} \\
\cmidrule(lr){3-6}
\cmidrule(lr){7-10}
\cmidrule(lr){11-12}
& & 
\textbf{Mod.} & \textbf{Dom.} & \textbf{Res.} & \textbf{Sta.}
& \textbf{Host} & \textbf{Path.} & \textbf{OrgL.} & \textbf{Org.}
& \textbf{Macro} & \textbf{Micro} \\
\midrule

\multirow{2}{*}{Gemini 2.5-Pro~\cite{comanici2025gemini2.5}} 
& \ding{55} 
& \textbf{99.56} & \textbf{63.86} & \textbf{80.25} & 59.02 
& \textbf{46.82} & \textbf{33.19} & 53.89 & 66.91 
& 62.93 & 65.07 \\

& \ding{51}
& 98.80 & 55.69 & 66.66 &\textbf{62.14} 
& 35.31 & 32.89 & \textbf{59.81} & \textbf{79.01} 
& 61.28 & 61.44 \\


\midrule

\multirow{2}{*}{QwenVLM-7B~\cite{bai2023qwen}} 
& \ding{55} 
& 68.05 & 52.11 & 82.72 & 49.68 
& 30.75 & 25.21 & 43.15 & 64.20 
& 51.98 & 48.58 \\

& \ding{51}
& \textbf{91.92} & \textbf{57.28} & \textbf{85.19} & \textbf{50.52} 
& \textbf{37.47} & \textbf{25.26} & \textbf{49.63} & \textbf{66.67} 
& \textbf{57.99} & \textbf{57.72} \\


\bottomrule
\end{tabular}
\vspace{-1em}
\label{tab:cot_results}
\end{table*}

\begin{table*}[t]
\centering
\caption{Few-shot in-context calibration results on counting and localization tasks.}
\label{tab:incontext_calibration}
\begin{tabular}{l|c|c|cc|c|cc}
\toprule
\multirow{2}{*}{\textbf{Model}} 
& \multirow{2}{*}{\textbf{Setting}} 
& \multicolumn{3}{c|}{\textbf{Counting}} 
& \multicolumn{3}{c}{\textbf{Localization}} \\

\cmidrule(lr){3-5} \cmidrule(lr){6-8}

& 
& \textbf{MC Acc}$\uparrow$
& \textbf{MAE}$\downarrow$ 
& \textbf{RMSE}$\downarrow$ 
& \textbf{MC Acc}$\uparrow$
& \textbf{mAP@0.5}$\uparrow$ 
& \textbf{IoU}$\uparrow$ \\

\midrule

GPT-5~\cite{OpenAI2025GPT5} & 0-shot & \textbf{71.34} & \textbf{2.07} & \textbf{3.02} & \textbf{83.21}& \textbf{56.66} & \textbf{50.53} \\
GPT-5~\cite{OpenAI2025GPT5} & 1-shot & 68.29 & 2.28 & 4.13 &68.73& 53.73 & 43.34 \\

\midrule

QwenVLM-7B~\cite{bai2023qwen} & 0-shot &46.45& 3.38 & 9.24 &\textbf{89.44}& \textbf{65.05} & \textbf{68.64} \\
QwenVLM-7B~\cite{bai2023qwen} & 1-shot &\textbf{47.18}& \textbf{3.11} & \textbf{6.73} &72.76& 62.79 & 65.85 \\

\bottomrule
\end{tabular}
\end{table*}

\noindent \textbf{Counting and Localization Performance.}
In addition to recognition tasks, we further evaluate VLMs on counting and localization under two evaluation settings: multiple-choice selection and direct prediction. In the multiple-choice setting, models select the correct answer from candidate options. In the direct prediction setting, models directly generate the target count or bounding box coordinates, which better reflects their ability to perform numerical reasoning and spatial grounding without predefined options. The results are presented in Table~\ref{tab:regression_results}.
From the multiple-choice results, closed-source models generally achieve higher counting accuracy than open-source models, suggesting stronger reasoning ability when candidate options are provided. GPT-5 achieves the highest multiple-choice counting accuracy, while QwenVLM-7B performs best among open-source models.
In contrast, direct prediction reveals larger performance gaps. For counting, GPT-5 achieves the lowest MAE and RMSE, indicating more precise numerical estimation, whereas most open-source models exhibit substantially larger deviations from the ground truth. For localization, closed-source models do not consistently outperform open-source counterparts. The Gemini series exhibits weak performance, whereas QwenVLM-7B surpasses GPT-5 and achieves the best spatial grounding results. 
To further analyze this difference in spatial grounding ability, we visualize representative localization results from GPT-5 and QwenVLM-7B in Fig.~\ref{fig:loc_vis}.
QwenVLM-7B produces bounding boxes that more tightly align with the target objects, indicating stronger spatial grounding. In contrast, GPT-5 roughly identifies object locations but produces less precise bounding boxes, while appearing less sensitive to background noise.


\subsection{Performance across Domains and Modalities}
To assess VLM robustness, we evaluate 4 representative models: GPT-5, Gemini 2.5-Pro, LLaVA-13B and QwenVLM-7B, across different biological domains. As shown in Figure~\ref{fig:domain}, all models perform substantially better on nematology and cell biology, while botany and mycology remain challenging. Specifically, GPT-5 attains the highest accuracy on nematology and botany, followed by Gemini 2.5-Pro, whereas LLaVA-13B and QwenVLM-7B lag notably behind. Gemini 2.5-Pro consistently surpasses QwenVLM-7B and LLaVA-13B across all domains but remains slightly weaker than GPT-5 on nematology. In contrast, GPT-5 exhibits the lowest accuracy on cell biology among all models. 
Figure~\ref{fig:modality} further summarizes performance across imaging modalities. Most models achieve their highest accuracy on electron micrographs, as the strong performance on electron images may be attributed to their more uniform and structured appearance. Fluorescence images provide the second-best results, and light microscopy generally yields the lowest overall performance. However, GPT-5 shows a different pattern: it achieves the best performance on light microscopy, which is the most challenging modality for open-source VLMs. This suggests that GPT-5 is better adapted to light microscopy images.



\begin{figure*}[t]
    \centering
    \begin{subfigure}[t]{0.38\textwidth}
        \centering
        \includegraphics[width=\linewidth]{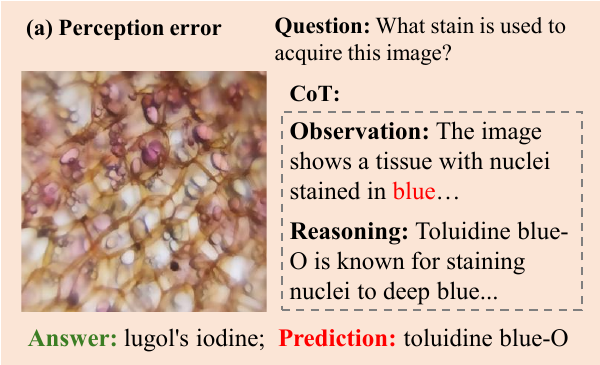}
        \caption{Perception error.}
        \label{fig:error_ex_1}
    \end{subfigure}
    \begin{subfigure}[t]{0.38\textwidth}
        \centering
        \includegraphics[width=\linewidth]{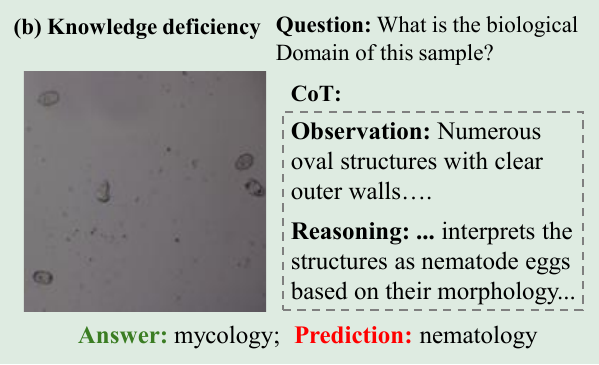}
        \caption{Knowledge deficiency.}
        \label{fig:error_ex_2}
    \end{subfigure}
    \begin{subfigure}[t]{0.22\textwidth}
        \centering
        \includegraphics[width=\linewidth]{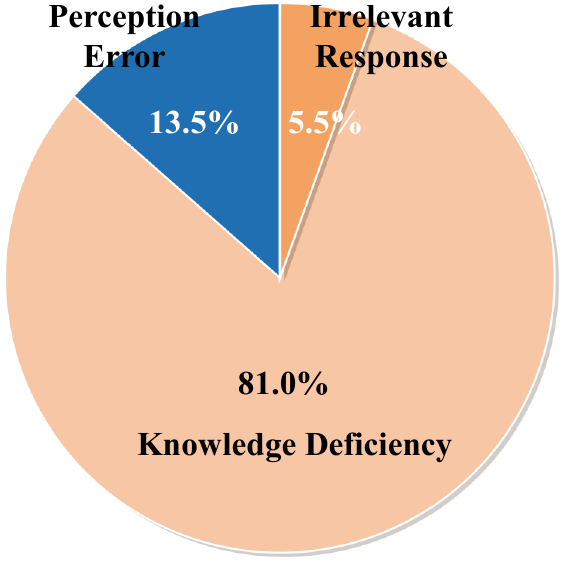}
        \caption{Error types.}
        \label{fig:error_dist}
    \end{subfigure}
    \caption{\textbf{Error analysis.} (a) Example of perception errors. (b) Example of knowledge deficiencies. (c) Distribution of error types.}
    \label{fig:error_analysis}
\end{figure*}

\begin{table*}[t]

\caption{Impact of RAG on microscopic VQA tasks. $\dagger$ denotes methods using RAG.}
\centering
\begin{tabular}{l|cccc|cccc|cc}
\toprule
\multirow{2}{*}{\textbf{Model}} &
\multicolumn{4}{c|}{\textbf{Image-Level}} &
\multicolumn{4}{c|}{\textbf{Biological-Level}} &
\multicolumn{2}{c}{\textbf{Overall}} \\
\cmidrule(lr){2-5}
\cmidrule(lr){6-9}
\cmidrule(lr){10-11}
& \textbf{Mod.} & \textbf{Dom.} & \textbf{Res.} & \textbf{Sta.}
& \textbf{Host} & \textbf{Path.} & \textbf{OrgL.} & \textbf{Org.}
& \textbf{Macro} & \textbf{Micro} \\
\midrule

{LLaVA-13B~\cite{liu2024llava}}
& 55.50 & 44.68 & 35.80 & 39.30
& 32.43 & 28.32 & 37.22 & 24.69
& 37.24 & 41.88 \\

{LLaVA-13B$^\dagger$~\cite{liu2024llava}}
& \textbf{86.17} & \textbf{52.73} & \textbf{85.19} & \textbf{52.66}
& \textbf{54.57} & \textbf{55.43} & \textbf{43.15} & \textbf{32.10}
& \textbf{57.75} & \textbf{60.21} \\

\midrule

{LLaVA-Next-13B~\cite{li2024llavanext}}
& 49.15 & 32.56 & 49.38 & \textbf{49.97}
& 40.10 & 29.56 & 32.04 & 46.91
& 41.20 & 39.63 \\

{LLaVA-Next-13B$^\dagger$~\cite{li2024llavanext}}
& \textbf{90.28} & \textbf{48.90} & \textbf{80.25} & {44.49}
& \textbf{50.91} & \textbf{57.51} & \textbf{36.30} & \textbf{56.79}
& \textbf{58.18} & \textbf{60.05} \\

\midrule

{QwenVLM-7B~\cite{bai2023qwen}}
& 68.05 & 52.11 & 82.72 & 49.68
& 30.75 & 25.21 & 43.15 & 64.20
& 51.98 & 48.58 \\

{QwenVLM-7B$^\dagger$~\cite{bai2023qwen}}
& \textbf{98.15} & \textbf{58.59} & \textbf{83.95} & \textbf{81.42}
& \textbf{56.75} & \textbf{39.69} & \textbf{62.08} & \textbf{68.41}
& \textbf{68.63} & \textbf{69.04} \\

\bottomrule
\end{tabular}
\label{tab:rag}
\end{table*}

\subsection{Reasoning and In-Context Prompting}

\noindent \textbf{Effect of CoT prompting.}
Chain-of-Thought (CoT) prompting~\cite{wei2022cot} has shown strong effectiveness in arithmetic, commonsense, and symbolic reasoning tasks. To examine whether explicit reasoning guidance can similarly benefit microscopic understanding, we evaluate the impact of CoT prompting on identification tasks.
Specifically, we adopt a three-step CoT setting following~\cite{kojima2022cot_zero_shot}, where a structured prompting scheme instructs the model to follow an observation–reasoning–selection workflow when answering each question.
We conduct experiments using Gemini 2.5-Pro and QwenVLM-7B, the best-performing closed-source and open-source models in multiple-choice identification tasks, to ensure a stable and representative evaluation. Results with and without CoT are summarized in Table~\ref{tab:cot_results}.
Under the CoT setting, Gemini 2.5-Pro shows slight improvements on the Stain and Organ tasks but exhibits performance drops on most other tasks, leading to a lower overall accuracy. In contrast, QwenVLM-7B achieves noticeable gains across multiple tasks, suggesting that structured reasoning guidance may better support open-source models, which tend to rely more on explicit reasoning steps when handling microscopic cues.

\vspace{0.5em}

\noindent \textbf{Impact of In-Context Prompting.}
We evaluate the effect of in-context prompting in microscopic scenarios using a 1-shot setting with the best-performing models, GPT-5 and QwenVLM-7B, on both counting and localization tasks. Results in Table~\ref{tab:incontext_calibration} show that introducing a single in-context example does not lead to consistent performance improvements. Specifically, QwenVLM-7B shows a slight improvement in counting but degraded performance on localization, while GPT-5 exhibits performance drops on both tasks.
We attribute this degradation to misalignment between the in-context demonstration and the target image. Due to the high visual similarity, repetitive structures, and dense object distributions in microscopic images, models may over-rely on patterns from the demonstration rather than grounding predictions on the input image itself.

\subsection{Error Analysis}
As VLMs exhibit limited capability on PlantMicro, we conduct an error analysis to identify their major failure modes. For each benchmark task, we randomly select 25 erroneous samples and trace the CoT reasoning process of Gemini 2.5-Pro to determine at which stage the model deviates from the correct reasoning path.
The observed errors can be grouped into three categories: perception errors, knowledge deficiencies, and irrelevant responses.
In perception errors, the model fails at the earliest stage of CoT reasoning: visual observation.
Figure~\ref{fig:error_ex_1} presents an example where the stain color is misidentified as blue, which subsequently leads the model to predict toluidine blue O instead of the ground-truth label.
In contrast, knowledge deficiencies arise when the model correctly perceives the visual evidence but lacks the biological knowledge needed for accurate interpretation.
As demonstrated in Figure~\ref{fig:error_ex_2}, although the model correctly identifies the oval structure of spores, it mistakenly links them to nematode eggs, indicating that the model has difficulty distinguishing entities at a fine-grained level.
These two error types correspond to the observation and reasoning stages in the CoT process, respectively.
Irrelevant responses, by contrast, refer to cases in which the model produces semantically unrelated outputs.
According to the error distribution shown in Figure~\ref{fig:error_dist}, knowledge deficiency accounts for the majority of failures, indicating that current VLMs have not yet been effectively adapted to the plant microscopy domain.



\subsection{Investigation on RAG}
To examine whether domain-specific knowledge can improve VLM performance in microscopic scenarios, we introduce a retrieval-augmented generation (RAG) setting for open-source VLMs (marked with $^\dagger$). Specifically, we retrieve the most visually similar example from the benchmark pool based on visual embedding similarity and prepend it as an in-context demonstration. To ensure task consistency, retrieval is restricted to samples within the same task category. As shown in Table~\ref{tab:rag}, incorporating RAG consistently improves performance across most microscopic VQA tasks.
These results suggest that the performance bottlenecks of current VLMs largely stem from insufficient domain-specific knowledge.


\section{Conclusion}
In this paper, we introduce PlantMicro, a comprehensive benchmark for evaluating vision–language models in microscopic plant image understanding. By integrating diverse microscopy datasets spanning multiple biological domains and imaging modalities, PlantMicro enables systematic evaluation across recognition, counting, and localization tasks. 
Extensive experiments demonstrate that while state-of-the-art VLMs such as Gemini 2.5 and GPT-5 can achieve strong performance in image-level modality identification, they exhibit substantial degradation in fine-grained biological entity recognition, spatial localization, and object counting. 
Our error analysis indicates that these limitations primarily arise from insufficient domain-specific knowledge in plant microscopy. Further experiments show that incorporating retrieval-augmented generation (RAG) consistently improves performance, highlighting the potential of knowledge-enhanced approaches for microscopic visual understanding.
Overall, PlantMicro bridges the gap between general-purpose VLMs and domain-specific microscopic understanding, providing a foundation for future research in plant microscopy.

\section*{Acknowledgement}
This work was partially supported by the Grains Research and Development Corporation (GRDC) through the Analytics for the Australian Grains Industry (AAGI) Strategic Partnership (UOQ2301-010OPX) and the project Exploring Data Efficient Machine Learning for Australian Grains Disease Recognition (USQ2605-002BGX), and by The University of Queensland (DVCR2201A).

\bibliographystyle{ieeenat_fullname}
\bibliography{main}
\end{document}